\documentclass[journal]{IEEEtran}

\ifCLASSINFOpdf
\else
\fi

\usepackage[ruled, linesnumbered]{algorithm2e}
\usepackage{algorithmicx}
\usepackage{algpseudocode}
\usepackage{color}
\usepackage{caption}
\usepackage{pifont}
\usepackage{times}
\usepackage{epsfig}
\usepackage{graphicx}
\usepackage{amsmath}
\usepackage{amssymb}
\usepackage{yhmath}
\usepackage{epsfig}
\usepackage{graphicx}
\usepackage{amsmath}
\usepackage{amssymb}
\usepackage{bm}
\usepackage{algorithm2e}
\usepackage{multirow}
\usepackage{subfigure}
\usepackage{amsthm}
\usepackage{booktabs}
\usepackage{tabularx}
\usepackage{url}
\usepackage{graphicx}
\usepackage{subfigure} 
\usepackage{mathrsfs}
\usepackage{amsmath}
\usepackage{multirow}
\usepackage{rotfloat}
\usepackage{booktabs}
\usepackage{graphicx}
\usepackage{bbding}

\usepackage{makecell}

\usepackage{booktabs}
\usepackage{bbding}
\usepackage{multirow}
\usepackage{pifont}

\usepackage{bm}
\usepackage{booktabs}
\usepackage{color}
\usepackage{rotfloat}
\usepackage{tabularx}
\usepackage[misc]{ifsym}

\usepackage{multirow}
\usepackage{float}

\usepackage{rotfloat}
\usepackage{tabularx}
\usepackage[misc]{ifsym}

\usepackage{multirow}
\usepackage{float}

\usepackage{graphicx}
\usepackage{makecell}

\usepackage{bm}
\usepackage{booktabs}
\usepackage{color}
\usepackage{multirow}
\usepackage{rotfloat}
\usepackage{tabularx}
\usepackage[misc]{ifsym}

\usepackage{multirow}
\usepackage{float}

\usepackage{arydshln}

\begin{document}

\title{Joint Perceptual Learning for Enhancement and Object Detection in Underwater Scenarios}

\author{ Chenping Fu, Wanqi Yuan, Jiewen Xiao, Risheng Liu, \textit{Member, IEEE}, and Xin Fan, \textit{Senior Member, IEEE}

	\thanks{
		
		Xin Fan and Risheng Liu are with the DUT-RU International School of Information Science, Dalian University of Technology, Dalian 116024, China (Corresponding author: Xin Fan. e-mail: xin.fan@ieee.org).
		Chenping Fu is with the School of Software Technology, Dalian University of Technology, 116024, China.
     Wanqi Yuan and Jiewen Xiao are with the International School of Information Science and Engineering, Dalian University of Technology, Dalian, China.
}	  
   
}

\markboth{Journal of \LaTeX\ Class Files,~Vol.~14, No.~8, August~2015}
{Shell \MakeLowercase{\textit{et al.}}: Bare Demo of IEEEtran.cls for IEEE Journals}

\maketitle

\begin{abstract}
Underwater degraded images greatly challenge existing algorithms to detect objects of interest. Recently, researchers attempt to adopt attention mechanisms or composite connections for improving the feature representation of detectors. However, this solution does \textit{not} eliminate the impact of degradation on image content such as color and texture, achieving minimal improvements. Another feasible solution for underwater object detection is to develop sophisticated deep architectures in order to enhance image quality or features. Nevertheless, the visually appealing output of these enhancement modules do \textit{not} necessarily generate high accuracy for deep detectors. More recently, some multi-task learning methods jointly learn underwater detection and image enhancement, accessing promising improvements. Typically, these methods invoke huge architecture and expensive computations, rendering inefficient inference. Definitely, underwater object detection and image enhancement are two interrelated tasks. Leveraging information coming from the two tasks can benefit each task. Based on these factual opinions, we propose a bilevel optimization formulation for jointly learning underwater object detection and image enhancement, and then unroll to a dual perception network (DPNet) for the two tasks. DPNet with one shared module and two task subnets learns from the two different tasks, seeking a shared representation. The shared representation provides more structural details for image enhancement and rich content information for object detection. Finally, we derive a cooperative training strategy to optimize parameters for DPNet. Extensive experiments on real-world and synthetic underwater datasets demonstrate that our method outputs visually favoring images and higher detection accuracy.              
\end{abstract}   
     

\IEEEpeerreviewmaketitle

\section{Introduction}

\begin{figure*}
	\centering
	\includegraphics[width=7.2in,height=3.5in]{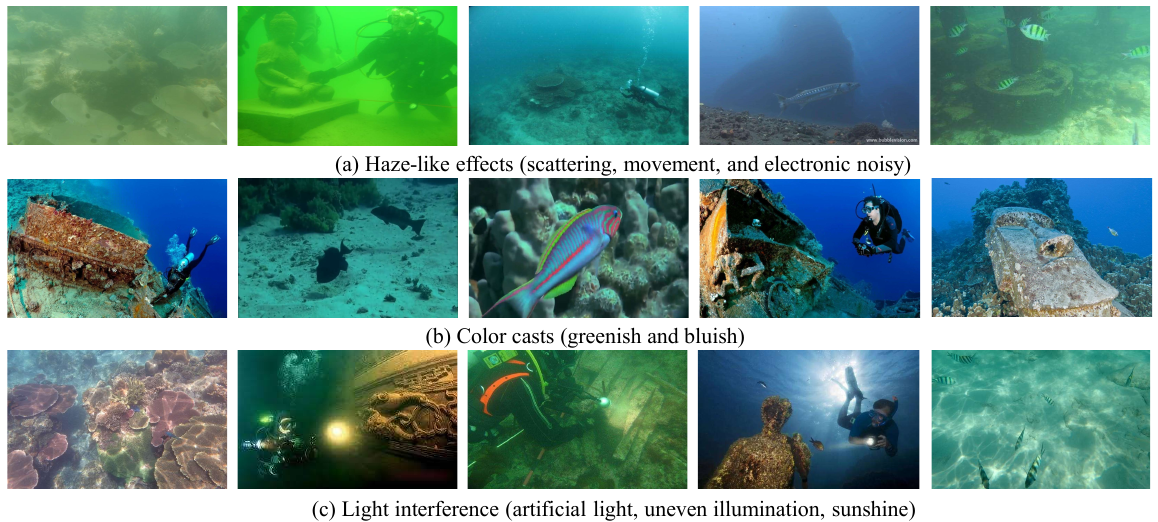}
	\caption{Examples of underwater environmental degradations include haze-like effects, color casts, and light interference. Images are selected from UEDB.}
	\label{show}
\end{figure*}

\IEEEPARstart{D}etecting objects of interest under underwater conditions are extremely important for applications such as marine monitoring and aquaculture \cite{chen1,minjun,chenlong,UDD}. However, underwater object detection suffers from low detection accuracy due to various environmental degradations (as shown in Fig.~\ref{show}). First are haze-like effects. The water medium and sediments scatter the light causing low-contrast and haze-like phenomena in underwater photography. The motion of electronics and objects can also produce blurring during high-dimensional imaging. Second are color distortions. Wavelength absorption usually causes a color reduction in the captured image, which leads to bluish or greenish underwater images \cite{minjun}. Third is light interference. Artificial lighting is widely used for underwater photography, and this non-uniform lighting causes vignetting in captured images \cite{light,shine}. Furthermore, the flickering affects always exist on sunshine days. This will cause the captured images with strong highlights in the shallow ocean \cite{shine}. These environmental degradations greatly impact the imaging content, which makes it difficult to recognize the object from the background \cite{cat}.

In practice, one solution for underwater object detection is to retrain generic detectors \cite{Cascade,Grid,guide,DETR,Swin,PVT} designed on high-quality image dataset (\textit{e.g.,} Pascal VOC \cite{VOC} and COCO \cite{COCO}). However, they suffer disappointed detection performance due to their poor feature representation in adverse conditions. Researchers thus improve the feature representation capacity of detection networks through operations such as composite connection \cite{UDD,FERNet} or attention strategy \cite{ACM}. However, this strategy does \textit{not} eliminate the impact of degradation on image content such as color and texture, achieving minimal improvements.    

Another feasible solution for underwater detection is to develop sophisticated deep architectures to enhance image/feature quality. For example, Sea-Thru \cite{sea-thru}, USICR \cite{USICR}, and FUnIE-GAN \cite{FUnIE-GAN} present enhancement algorithms to recover the color or texture of underwater images. Directly taking these enhancement networks as processing steps for improving image quality involves complicated architectures. Moreover, the visually appealing outputs of these enhancement modules do \textit{not} necessarily generate high accuracy for deep detection algorithms \cite{chenlong,chen1}.

Recently, multi-task learning has been introduced and applied well for computer vision fields. These methods aim to leverage information coming from related tasks so that each task can gain benefit \cite{bao2022generative,Carl}. Specifically, MGM \cite{bao2022generative}, MultiNet \cite{MultiNet}, and UWL \cite{UWL} explore multi-task learning within the setting of visual scene understanding in computer vision. There are also some multi-task works for joint learning underwater object detection and image enhancement, prompting enhancement modules to generate detection-favored and visually appealing images \cite{chenlong}. Typically, these methods invoke huge architectures and inefficient inference, hindering their applications to time-critical scenarios. Therefore, problem definitions and effective architecture designs are still challenges to jointly learning object detection and image enhancement in underwater scenes.  

\begin{figure*}[t]
	\centering
	\includegraphics[width=7.2in,height=3.2in]{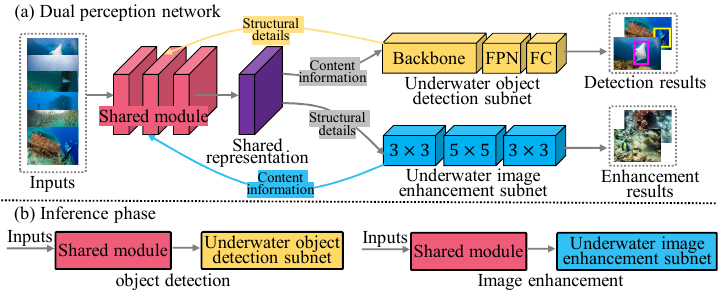}
	\caption{Workflow of our method. (a) The architecture of our dual perception network (DPNet). (b) During the inference phase, we only retain the shared module and a detection/enhancement subnet for detection/enhancement tasks. }
	\label{liucheng}
\end{figure*}

Indeed, underwater object detection and image enhancement are two complementary tasks. The shared representation learned from the two tasks can help each task. Specifically, the shared representation can provide more content information including color or texture details for underwater object detection. Simultaneously, the shared representation can provide more structural details for image enhancement. Motivated by these factual opinions, this paper proposes a bilevel optimization formulation for jointly learning underwater object detection and image enhancement, aiming to learn the shared representation that benefits the two different tasks and thereby improves the performance of the two tasks. This formulation unrolls to a dual perception network (DPNet), composed of one shared module and two task subnets. One subnet achieves detection tasks and the other one achieves enhancement tasks. The shared module links the two subnets and provides the shared representation for them. We also derive a cooperative training strategy to learn optimal parameters for DPNet. Fig.~\ref{liucheng} demonstrates our methodology framework. Our contributions are four-fold: 

\begin{itemize}
	\item We supply a general bilevel modeling perspective for jointly learning underwater object detection and image enhancement. It can accurately depict the latent correspondence between the shared representation, detection, and enhancement.   

	\item We devise a dual perception network for the bilevel model. DPNet \lq\lq{}seeks the shared representation from differences\rq\rq{} that provides rich content information for underwater object detection and more structural details for image enhancement.

	\item We derive a cooperative training scheme from the bilevel optimization formulation, yielding optimal network parameters for DPNet.

\end{itemize}

To prove the detection and enhancement efficiency of our approach, we evaluate the proposed model on synthetic and natural underwater datasets. Quantitative and qualitative evaluation results show that our proposed approach surpasses the accuracy of current state-of-the-art object detection and image enhancement methods.  

The rest of this paper is organized as follows: Sec.\ref{Relate} provides related works including underwater object detection and multi-task learning. Sec.~\ref{proposed} describes the proposed object detection method, which is used in underwater images. Sec.\ref{experiment} presents the comprehensive experimental results of our method compared with other methods. Sec.\ref{con} concludes our work.

\section{Related Works}\label{Relate}
\subsection{Underwater Object Detection}

Underwater object detection aims at determining what and where an object is in an underwater image. Compared with high-quality images, underwater images captured in complex environments confront evident quality degradation. Generic deep detectors \cite{yolo,SSD,Focal,fsaf,free,FoveaBox} designed on high-quality images thus have poor performance when retrained on underwater images. 

Some researchers attempt to improve the feature representation capacity of generic deep detectors \cite{UDD,FERNet,ACM}. For example, AquaNet \cite{UDD} designs two efficient components MFF and MBP using multi-scale features fusion and anti-aliasing operations. FERNet \cite{FERNet} proposes a composite connection backbone and anchor refinement scheme to boost the feature representation. CSAM \cite{ACM} proposes a novel channel sharpening attention module to fuse high-level image information for better feature utilization. SSoB \cite{yuan} proposes a mixed anti-aliasing block and resorts neural architecture search to build a detection backbone for underwater scenes. While promising, these methods do \textit{not} remove the influence of degradation factors on detection features, achieving limited improvement.     

Image enhancement algorithms can improve an underwater image to a sharp version.  Typically, Sea-Thru \cite{sea-thru} proposes a physical imaging model to recover lost colors in underwater images. FUnIE-GAN \cite{FUnIE-GAN} presents a conditional generative adversarial network-based model for recovering global content, color, and local texture of underwater images. USICR \cite{USICR} uses haze lines to solve color restoration problems in underwater images. Since these underwater enhancement algorithms are designed for visually appealing outputs, directly using them as processing steps for improving image quality for deep detectors may lead to suboptimal performance.

\subsection{Multi-task learning in the computer vision society}
Multi-task learning simultaneously learns multiple related tasks, aiming to be beneficial to each task. Many multi-task learning models have been proposed and applied in various deep learning applications in the computer vision society. For example, MGM \cite{bao2022generative} couples a discriminative multi-task network with a generative network for joint learning depth estimation, surface normal estimation, and semantic segmentation tasks. MultiNet \cite{MultiNet} uses a unified architecture to jointly learn classification, detection, and semantic segmentation tasks. UWL \cite{UWL} designs a principled approach to multi-task deep learning which weighs multiple loss functions by considering the homoscedastic uncertainty of each task.

Some researchers also attempt to design multi-task learning models for underwater enhancement and detection tasks, aiming to prompt the two tasks \cite{chenlong}. Specifically, HybridDetectionGAN \cite{chenlong} use the detection preceptor to provide feedback information in the form of gradients to guide the enhancement model, generating visually pleasing or detection favorable images. However, this manner involves a complicated architecture and occupies a huge inference time. In our paper, unlike the methods reported in \cite{chenlong}, our DPNet has a simple but efficient joint learning strategy and architecture. During the inference phase, we only retain the small shared model and a detection/enhancement subnet for completing detection/enhancement tasks.

\section{The Proposed Approach}\label{proposed}
Underwater imaging processes suffer from haze-like effects, color casts, and light interference. These evident quality degradations impact image content information including color and texture, which makes the detection difficult to distinguish objects from the background \cite{cat}.

Jointly learning underwater object detection and image enhancement benefits each of them, since they are two complementary works. Specifically, underwater image enhancement aims to recover the color/texture of images, which helps detectors to detect objects from harsh conditions. Underwater object detection attempts to decide what and where an object is in an image. The fundamental features of detection tasks contain rich structural information, which aids enhancement methods to make the image sharper. The shared representation between detection and enhancement thus benefits the two tasks. However, problem definitions and effective architecture designs are still challenges to this joint learning issue.

Inspired by these factual opinions, we introduce a general bilevel modeling perspective for learning a shared representation between underwater object detection and image enhancement. The shared representation provides content information for prompting detection and structural details for boosting enhancement. Then, we unroll the bilevel modeling to a dual perception network. Finally, we give a cooperative training scheme to learn optimal parameters for DPNet. Fig.\ref{liucheng} illustrates the idea, introduced next.

\subsection{Problem Formulation}
According to the general scheme of multi-task learning, we take a shared module to connect underwater detection and enhancement subnets. The shared module seeks the shared representation from the two subnets, providing more content information for detection through enhancement task properties and more structural details for enhancement through detection task capabilities. Therefore, its parameter optimization depends on the two subnets' optimization. Naturally, following the truism of Stackelberg's theory \cite{GAO,peter}, we formulate the joint problem of underwater object detection and image enhancement as a bilevel optimization model:   

\begin{equation}\label{e1}
\begin{aligned}
&\min_{\bm u}\;\mathcal{L}^{val}(\bm{\omega}^*, \Phi (\bm {u}))\\
&s.t.,\;\;\bm{\omega}^*=\arg\min_{\bm{\omega}}\mathcal{L}^{tr}(\bm{\omega}, \Phi (\bm u)),
\end{aligned}
\end{equation}
where $\Phi$ denotes the shared module with learnable parameters $\bm u$. $\mathcal{L}$ is the loss function of the joint problem, which is defined as the sum of the detection loss $\mathcal{L}_{det}$ and enhancement loss $\mathcal{L}_{enh}$. $\mathcal{L}^{val}$ denotes the loss with the validation set and $\mathcal{L}^{tr}$ denotes the loss with the training set.   
$\bm{\omega}=(\bm{\omega}_{det}, \bm{\omega}_{enh})$, where $\bm{\omega}_{det}$ and $\bm{\omega}_{enh}$ are parameters of the detection and enhancement subnets, respectively. 

The above bilevel modeling actually builds an explicit relationship among the shared module, detection, and enhancement subnets. The shared module parameter $\bm {u}$, the upper-level parameter, is updated along with lower-level parameters. $\bm{\omega}_{det}$ and $\bm{\omega}_{enh}$, the lower-level parameters, are separately optimized by the detection loss $\mathcal{L}_{det}$ and enhancement loss $\mathcal{L}_{enh}$ due to the two subnet independence. As is shown in Fig.~\ref{liucheng} (a), we unroll the bilevel optimization formulation to a novel DPNet to solve the problem in Eq.(\ref{e1}).

\subsection{Dual Perception Network}
Most multi-task learning methods for underwater enhancement and object detection involve a complicated architecture and occupy a huge inference time. To overcome these problems, we devise a simple but efficient joint learning architecture for the two tasks (\textit{i.e.,} DPNet). As is shown in Fig.~\ref{liucheng} (a), DPNet consists of three parts: a shared module, an object detection subnet, and an image enhancement subnet. During the training phase, the shared module, placed before the two subnets, is designed for providing shared representation containing more structure and content information (\textit{e.g.,} color or texture) to the two subnets. The enhancement and object detection subnets are trained for image enhancement and detection, respectively. As is shown in Fig.~\ref{liucheng} (b), we only retain the shared module and a detection/enhancement subnet for detection/enhancement tasks during the inference phase. This manner reduces architectural complexity, achieving efficient memory space and inference times. As is shown in Fig.~\ref{liucheng}, we introduce the DPLNet from its three components.   

\textbf{The shared module} extracts the shared representation of the input image that contains important information for simultaneous benefiting underwater enhancement and detection. For one thing, the shared representation contains much useful content information including recovered color and texture, which is beneficial for detectors to detect objects from degraded environments. For another thing, the shared representation contains much structural information, which is beneficial for enhancement methods to generate sharper images.      

How to design a suitable structure for the shared module plays a vital role in implementation. First, underwater missions often deploy to mobile CPUs, requiring lightweight modules. Second, the shared module should contain more low-level information including color, texture, and structure, as it is applied to provide shared representation for detection and enhancement subnets. For these considerations, we construct the shared module with few layers and small kernels. Specifically, the shared module contains three convolution layers with kernels of $3\times3$, $5\times5$, and $3\times3$, respectively. The channels of the three convolution layers are set as 32. 

\textbf{Underwater image enhancement subnet} is responsible for improving the visualization quality of underwater images. For image enhancement tasks, downsampling and upsampling operations tend to be used together, since downsampling operations cause some problems including missing image information. This manner invokes complex architectures, and we thus design the underwater image enhancement subnet with a plain convolution without downsampling. The subnet consists of three $3\times 3$ convolutional layers with stride 1 and padding 1. The first 2 layers are followed by a batch normalization layer and Relu activation function. The last layer is followed by a batch normalization layer and Sigmoid activation function. The sigmoid activation function is used to make the output in [0,1]. The subnet produces an image with $H\times W\times 3$, where $H$, $W$, and $3$ are the height, weight, and channels of the image. 

\textbf{Underwater object detection subnet} is responsible for classification and regression. Here, we adopt RetinaNet \cite{RetinaNet} as the detection subnet. To adapt the shared module, the channel of the first layer of RetinaNet is set to 32. There are two reasons for us to choose RetinaNet as our backbone. For one thing, RetinaNet proposes the focal loss that focuses on learning hard examples to realize a balance between classes during training. For another thing, RetinaNet employs a feature pyramid network (FPN) \cite{FPN} that provides a top-down pathway and lateral connections. RetinaNet thus enables the construction of higher resolution layers from a rich semantic layer, which can significantly improve detection accuracy in harsh conditions. For these reasons, RetinaNet has the potential capacity of coping with the challenge of detecting objects in poor visibility conditions.

\begin{figure*}[t]
	\centering
	\includegraphics[width=\textwidth,height=2in]{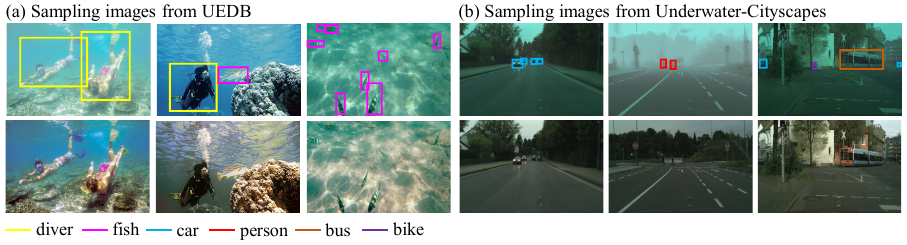}
	\caption{Sampling images from UEDB and Underwater-Cityscapes. Top row: raw underwater images with semantic labels; Bottom row: the corresponding reference images.}
	\label{data}
\end{figure*}

\subsection{Cooperative Training Strategy}       
The bilevel optimization naturally derives a cooperative training strategy to obtain optimal network parameters $\bm {u}$ and $\bm{\omega}$. Let $t \in [1, T]$ be the iteration times. The loss gradients with respect to $\bm {u}$ is calculated as:

\begin{equation}
\label{eq:gradient}
\begin{aligned}
\frac{\partial \mathcal{L}^{val}}{\partial \bm {u}}&= \frac{\partial \mathcal{L}^{val}(\bm{\omega}^t,\Phi(\bm {u}))}{\partial \Phi(\bm {u})} \frac{\partial \Phi(\bm {u})}{\partial \bm {u}} + \frac{\partial \mathcal{L}^{val}(\bm{\omega}^t, \Phi (\bm {u}))}{ \partial \bm{\omega}^t} \frac{ \partial \bm{\omega}^t}{\partial \bm {u}}
\\&= \frac{\partial \mathcal{L}^{val}(\bm{\omega}^t,\Phi(\bm {u}))}{\partial \Phi(\bm {u})} \frac{\partial \Phi(\bm {u})}{\partial \bm {u}} 
\\& + \frac{ \partial \mathcal{L}^{val}(\bm{\omega}^t,\Phi(\bm {u}))}{\partial \bm{\omega}^t} \frac{\bm \partial (\bm{\omega}^{t-1}-\frac{\partial \mathcal{L}^{tr}(\bm{\omega}^{t-1}, \Phi(\bm {u}))}{\partial \bm{\omega}})}{\partial \bm {u}}
\end{aligned}
\end{equation}

The Eq.~\ref{eq:gradient} reveals that the shared module, underwater image enhancement subnet, and object detection subnet are simultaneously updated. After the training process, the shared module trained with this strategy can generate a shared representation that is conducive to enhancement and detection, since the shared module is simultaneously optimized by the two tasks. Meanwhile, the detection and enhancement subnets can achieve their tasks, respectively.

\begin{table*}[h]
\centering

\renewcommand{\arraystretch}{1.3}
\setlength{\tabcolsep}{3.1mm}{
          \begin{tabular}{|c|cc|c|cccccccc|c|}
               \hline
               \multirow{2}{*}{Methods}& \multicolumn{3}{c|}{UEDB dataset}& \multicolumn{9}{c|}{Underwater-Cityscapes dataset}\\
               \cline{2-13}   
               & Diver &Fish&mAP &Person &Rider&Car&Truck&Bus&Train&Moto&Bike&mAP\\

\hline
\textbf{General detectors:}& & & &&&&&&&&&\\

 Free Anchor & 26.1 &25.0 &25.6 &33.9&29.7&55.0&22.8&28.6&26.7&21.3&27.7&30.7\\

               Fovea Box & 27.3 &23.0 &25.2 &31.2&26.3&50.8&23.8&31.6&25.1&27.5&25.8&30.3\\

				YOLOX& 26.2 &20.9 &23.6&34.1&27.9&55.6&25.7&44.2&29.4&28.4&26.1&33.9\\

RetinaNet& 32.4 &35.6 &34.0&34.4&34.8&59.2&24.0&43.1&27.3&30.9&32.3&35.8\\

              Grid RCNN & 26.5 &28.9 &27.7 &36.5&30.8&56.8&26.0&40.5&23.3&31.1&31.8&34.6\\

               PVTv1& 34.7 &28.6 &31.7 &34.4&30.2&54.8&26.7&36.8&22.1&27.4&30.6&32.9\\
               \hline
\textbf{Underwater detectors:}& & & &&&&&&&&&\\

CSAM& 35.3&35.2 &35.3&36.6&30.5&53.4&38.4&40.6&31.3&30.7&28.2&36.2\\
                AquaNet& 33.8 &34.8  &34.3 &34.6&28.2&54.4&25.2&31.3&28.8&30.5&34.6&33.5\\

			    FERNet& 34.0&38.6&36.3 &31.5&32.4&53.0&26.3&41.2&30.3&34.2&33.3&35.3\\
\hline

\textbf{Enhancement+detection:}& & & &&&&&&&&&\\

FUnIE-GAN+FERNet& 33.2 &35.5 &34.4 &30.0&30.8&51.4&24.2&37.0&25.4&31.8&32.7&32.9\\

FUnIE-GAN+RetinaNet& 30.2 &30.5 & 30.4&33.0&32.2&57.4&23.2&40.0&26.4&28.8&30.7&34.0\\

ERH+FERNet& 34.3 &35.2&34.8&35.1&31.5&50.3&23.4&39.2&27.5&32.3&29.6&33.6\\

ERH+RetinaNet& 30.3 &33.2&31.8&33.1&33.0&58.3&24.0&41.6&26.8&29.2&31.6&34.7\\

\hline

\textbf{Multi-task learning:}& & & &&&&&&&&&\\

HybridDGAN& 40.3 &39.3 &39.8&38.4&37.7&58.6&35.6&43.8&30.0&27.5&28.8&37.6\\
\hline
\hline

\textbf{Ours:}& & & &&&&&&&&&\\
               DPNet&\textbf{44.2} &\textbf{43.1} &\textbf{43.7} &\textbf{42.1}&\textbf{40.6}&\textbf{60.2}&\textbf{36.3}&\textbf{44.6}&\textbf{32.6}&\textbf{30.8}&\textbf{31.3}&\textbf{39.8}\\
               \hline

 \end{tabular}}

 \caption{Comparisons of the detection accuracy on UEDB and Underwater-Cityscapes datasets.}
\label{tab1}
\end{table*}

\section{Experiments}\label{experiment}
\subsection{Experimental Configurations}
In this section, we introduce experimental datasets including Underwater-Cityscapes and UEDB. Then, we brief implementation details of DPNet.

\textbf{Comparision dataset.} Some datasets have been published for underwater object detection. Typically, URPC 2018 $\thicksim$ 2022 \footnote{http://2022.urpc.org.cn/index.html} are created to support underwater robot professional contests, consisting of 4 categories. UDD \cite{UDD} is an underwater object-grabbing dataset, and images are manually collected by divers on Zhangzi Island, Dalian, China. RUOD \cite{RUOD} is an object detection dataset for general underwater scenes, containing various environmental challenges. Some underwater image enhancement datasets also have been published. Representatively, UIEB \cite{UIEB} includes 890 real-world underwater images and their corresponding reference images. Blasinski \textit{et al.} \cite{UISS} provides an open-source underwater image simulation tool and a three-parameter formation model. Duarte \textit{et al.} \cite{Duarte} simulates underwater image degradation using milk, chlorophyll, or green tea in a tank. However, most existing underwater image datasets only support one task and cannot support detection and enhancement at the same time.

To conduct our experiments, we compose two datasets by synthetic methods or annotation manners. The two datasets have semantic labels for detection and reference images for enhancement. Specifically, Cityscapes is a clear weather image set, which includes car, truck, motorcycle/bike, train, bus, rider, and person. As per \cite{UNP}, we simulate underwater scenes on Cityscapes \cite{Cityscapes} to synthesize the Underwater-Cityscapes. Therefore, Underwater-Cityscapes has the same labels as Cityscapes and takes Cityscapes as the reference images. Underwater-Cityscapes consists of a train/val set with 2975 images and a test set with 500 images. For further evaluating our DPNet on real-world underwater scenes, we also build a real-world underwater enhancement $\&$ detection benchmark called UEDB. UEDB is derived from a real-world underwater image enhancement dataset UIEB \cite{UIEB}. We first discard some images in UIEB according to two criteria: (1) An image without objects should be discarded. (2) An image with too many objects should be discarded due to difficulties in annotating them accurately and completely with confidence. We then annotate these retaining images to construct UEDB. UEDB includes two categories of diver and fish, containing 505 real-world underwater images and their corresponding reference images. Several sampling and the corresponding reference images from UEDB and Underwater-Cityscapes are presented in Fig.~\ref{data} (a)-(b). 

\textbf{Implement details}
We initialize the shared module and underwater image enhancement subnet with kaiming\_init. The underwater object detection subnet is initialized with weights pretrained on ImageNet \cite{ImageNet} (150 epochs). We train DPLNet for 24 epochs with Adam optimizer and batch size 2. The initial learning rate is 0.002 and exponential decay. We resize all image sizes to 800 $\times$ 1333 for UEDB and Underwater-Cityscapes as input. Our approach is implemented on PyTorch with an NVIDIA Tesla V100 GPU. 

\subsection{Underwater Object Detection Results}
We compare DPNet with popular object detection methods to evaluate the enhancement performance. There are four types of methods: general detectors that are originally designed on high-quality datasets, underwater detectors which improve the feature representation depending on the architecture strategy, enhancement+detection for cascading underwater image enhancement and detection, and multi-task based methods jointly learning underwater enhancement and detection. Here, we use common evaluation metrics AP (Average Precision) for each class and mAP (mean Average Precision) to measure the detection performance. 

\textbf{Compared with general detectors.} We compare our DPNet with general detectors on UEDB and Underwater-Cityscapes datasets, summarizing the comparisons in Tab~\ref{tab1}. We retrain these general detectors on the two underwater datasets, however, their performance on detection tasks is disappointing. DPNet achieves 43.7\% and 39.8\% mAP on UEDB and Underwater-Cityscapes datasets respectively, surpassing these methods by a large margin. For example, DPNet surpasses RetinaNet and PVTv1 by 9.7\% and 12.0\% mAP on the UEDB dataset, respectively. DPNet outperforms RetinaNet and Grid RCNN by 4.0\% and 5.2\% mAP on the Underwater-Cityscapes dataset, respectively.

\textbf{Compared with underwater detectors.}
As is shown in Tab.~\ref{tab1}, we compare our DPNet with representative underwater detection methods. CSAM \cite{ACM}, AquaNet \cite{UDD}, and FERNet \cite{FERNet} attempt to improve the feature representation capacity of detectors in underwater scenes through attention strategy and composite connection. They achieve limited gains since these methods do \textit{not} essentially remove the impact of degradation on underwater images. DPNet achieves a promising improvement. Specifically, DPNet surpasses CSAM, AquaNet, and FERNet by 8.4\%, 9.4\%, and 7.4\% mAP on UEDB, respectively. DPNet outperforms CSAM, AquaNet, and FERNet by 3.6\%, 6.3\%, and 4.5\% mAP on Underwater-Cityscapes, respectively.

\textbf{Compared with enhancement+detection.} We also compare the results of our DPNet with some enhancement+detection methods. Here, we use underwater image enhancement networks (\textit{i.e.,} FUnIE-GAN \cite{FUnIE-GAN} and ERH \cite{ERH}) as image processing steps and perform detection using RetinaNet and HybridDGAN trained on enhanced images. As is shown in Tab.~\ref{tab1}, we find that an underwater enhancement network as an image processing step is not always beneficial for underwater image detection. Similar phenomena are also reported in \cite{minjun,chen}. The phenomenon demonstrates that the visually appealing outputs of enhancement modules do \textit{not} necessarily generate high accuracy for deep detection algorithms. 
\begin{table}[t]
\centering
\renewcommand{\arraystretch}{1.4}
\setlength{\tabcolsep}{1mm}{
          \begin{tabular}{|c|ccc|ccc|}
               \hline
               \multirow{2}{*}{Methods}& \multicolumn{3}{c|}{UEDB dataset}& \multicolumn{3}{c|}{Underwater-Cityscapes dataset}\\
               \cline{2-7}   
               &SSIM $\uparrow$ &PSNR $\uparrow$ &PCQI $\uparrow$ &SSIM $\uparrow$ &PSNR $\uparrow$ &PCQI$\uparrow$\\

\hline

UDCP&0.553&12.554&0.532&0.689&22.435&0.887 \\

ERH&0.721&20.230&\textbf{0.862}&0.884&23.776&0.932\\

HybridDGAN&0.696&22.319&0.714&0.835&24.675&0.904\\

FUnIE-GAN &0.657&19.420&0.580&0.731&23.843&0.892\\

\hline
\hline

DPNet &\textbf{0.826}&\textbf{23.325}&0.856&\textbf{0.954}&\textbf{26.343}&\textbf{0.976}\\
               \hline

 \end{tabular}}

 \caption{Full-reference comparisons of the enhancement performance on UEDB and Underwater-Cityscapes datasets.}
\label{tab2}
\end{table}
\begin{table}[t]
\centering
\renewcommand{\arraystretch}{1.4}
\setlength{\tabcolsep}{2mm}{
          \begin{tabular}{|c|cc|cc|}
               \hline
               \multirow{2}{*}{Methods}& \multicolumn{2}{c|}{UEDB dataset}& \multicolumn{2}{c|}{Underwater-Cityscapes dataset}\\
               \cline{2-5}   
               &UIQM $\uparrow$ & NIQE $\downarrow$ & UIQM $\uparrow$ & NIQE $\downarrow$\\

\hline

UDCP&1.785&4.280&2.189&5.433\\

ERH&3.024&\textbf{4.054}&3.122&3.675\\

HybridDGAN&2.866&4.803&3.087&3.850\\

FUnIE-GAN &2.675&5.463&2.980&4.076\\

\hline
\hline

DPNet &\textbf{3.403}&4.189&\textbf{4.434}&\textbf{3.332}\\
               \hline

 \end{tabular}}

 \caption{Non-reference Comparisons of the enhancement performance on UEDB and Underwater-Cityscapes datasets.}
\label{tab3}
\end{table} 

\begin{figure*}[t]
	\centering
	\includegraphics[width=\textwidth,height=2.7in]{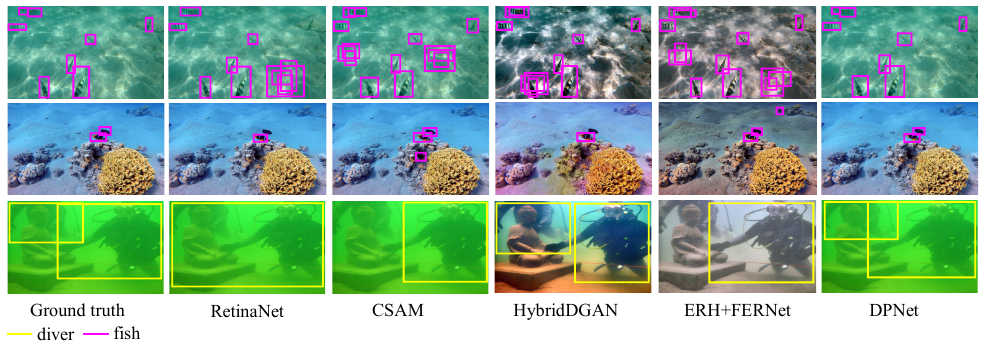}
	\caption{Visualization of detection results on UEDB dataset. The examples from top to bottom are light interference, color casts, and haze-like effects, respectively.}
	\label{de1}
\end{figure*}  
\begin{figure*}[t]
	\centering
	\includegraphics[width=\textwidth,height=2.1in]{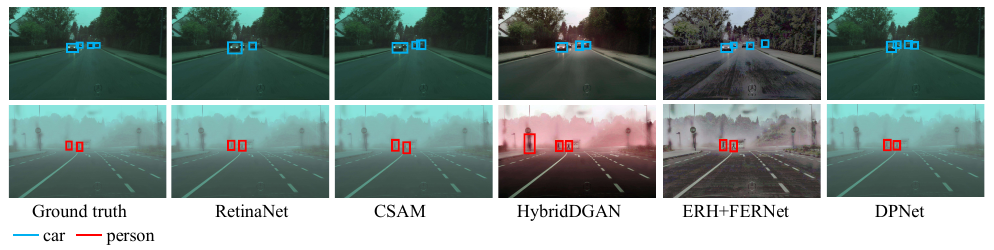}
	\caption{Visualization of detection results on Underwater-Cityscapes dataset. The examples from top to bottom are color casts and haze-like effects.}
	\label{de2}
\end{figure*}
\begin{table*}[h]
\centering
\renewcommand{\arraystretch}{1.2}
\setlength{\tabcolsep}{5.5mm}{
          \begin{tabular}{|c|cc|cc|cc|}
               \hline
               \multirow{2}{*}{Methods}& \multicolumn{2}{c|}{Underwater object detection}& \multicolumn{2}{c|}{Underwater image enhancement}&\multicolumn{2}{c|}{Underwater detection and enhacement}\\
               \cline{2-7}   
               &Size. (M)&FPS ($f\cdot s^{-1}$)&Size (M)&FPS ($f\cdot s^{-1}$)&Size (M)&FPS ($f\cdot s^{-1}$)\\

\hline

HybridDGAN&148.7&2.4&148.7&2.4&148.7&2.4\\

DPNet&70.3&4.3&30.2&5.5&85.0&3.5\\

Increase&78.4$\downarrow$&1.9$\uparrow$&118.5$\downarrow$&3.1$\uparrow$ &63.7$\downarrow$&1.1$\uparrow$\\

\hline
 \end{tabular}}
 \caption{Efficiency comparison on DPNet and HybridDGAN.}
\label{tab4}
\end{table*}

\begin{figure*}[h]
	\centering
	\includegraphics[width=\textwidth,height=4.7in]{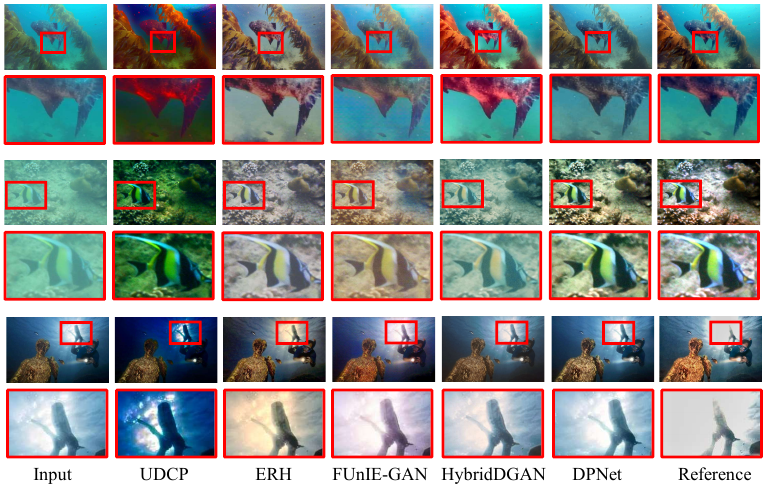}
	\caption{Visualization of enhancement results on Underwater-Cityscapes dataset. The examples from top to bottom are color casts, haze-like effects, and light interference, respectively.}
	\label{en1}
\end{figure*}  
\begin{figure*}[h]
	\centering
	\includegraphics[width=\textwidth,height=3.5in]{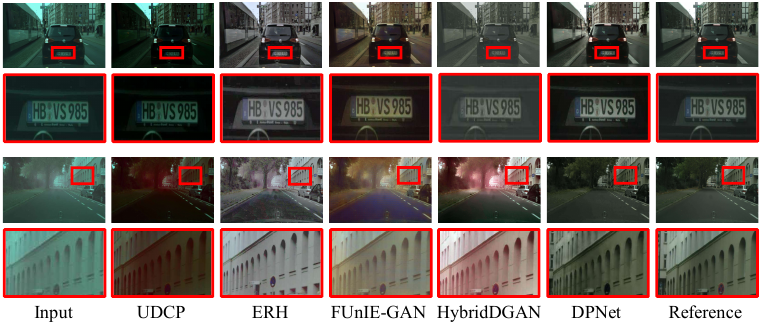}
	\caption{Visualization of enhancement results on UEDB dataset. The examples from top to bottom are color casts and haze-like effects, respectively.}
	\label{en2}
\end{figure*}

\textbf{Compared with multi-task methods.} HybridDGAN \cite{chenlong} uses hybrid underwater image synthesis and perceptual enhancement models to jointly learn underwater detection and enhancement. As is shown in Tab.~\ref{tab1}, HybridDGAN achieves more progress than CSAM, AquaNet, and FERNet. Our DPNet further improves the accuracy performance and achieves the best performance. Typically, DPNet outperforms HybridDGAN by 3.9\% and 2.2\% mAP on UEDB and Underwater-Cityscapes datasets, respectively.

\subsection{Performance Analysis}
In this part, we analyze our DPNet from image enhancement properties, efficiency performance, and environmental degradation performance.   

\textbf{Underwater image enhancement results.} We compare DPNet with popular underwater image enhancement methods to evaluate the enhancement performance. Here, we take three common full-reference metrics, including Structural Similarity (SSIM), Peak Signal-to-Noise Ratio (PSNR), and Patch-based Contrast Quality Index (PCQI) \cite{PCQI}. Higher PSNR, SSIM, and PCQI scores are closer to the reference image. Besides, for comprehensive comparison, we also employed two non-reference metrics, including Underwater Image Quality Measurement (UIQM) \cite{UIQM} and Natural Image Quality Evaluator (NIQE) \cite{NIQE}. The common underwater image quality measure UIQM consists of three attribute measures, namely, the underwater image colorfulness measure (UICM), the underwater image sharpness measure (UISM), and the underwater image contrast measure (UIConM). Higher UIQM scores mean higher image quality. NIQE is a popular non-reference image quality assessment method for evaluating the naturalness of a single image; a minor NIQE score means higher image quality.

As is shown in Tab.~\ref{tab2}, UDCP \cite{UDCP} is a physical model-based method, that achieves lower metric scores than other methods. ERH \cite{ERH} is the latest underwater vision reconstruction method, FUnIE-GAN \cite{FUnIE-GAN} is the representative data-driven method, and HybridDGAN \cite{chenlong} jointly learns enhancement and detection. ERH, FUnIE-GAN, and HybridDGAN, especially ERH, play better enhancement on UEDB and Underwater-Cityscapes datasets. Our DPNet achieves the best results on most metrics compared with these methods, especially in the synthetic dataset Underwater-Cityscapes. These experimental results demonstrate that DPNet can complete underwater image enhancement tasks well from the perspective of full-reference comparisons.

As is shown in Tab.~\ref{tab3}, the multi-task based HybridDGAN achieves acceptance performance. ERH plays better than most methods in general. Our DPNet wins on most non-reference metrics compared with all these methods. Typically, DPNet achieves the best results on the synthetic dataset Underwater-Cityscapes. These experimental results further show that DPNet can complete underwater image enhancement tasks well from the perspective of non-reference comparisons. 

\textbf{Efficiency analysis.} We compare the proposed methods in terms of speed and parameter size, as is shown in Tab.~\ref{tab4}. Compared to the multi-task learning method (\textit{i.e.,} HybridDGAN) for underwater detection and enhancement, DPNet has much fewer parameters and higher speed. Typically, our method can achieve 4.3 FPS, which is well-qualified for real-time detection tasks. DPNet has a higher enhancement speed compared to HybridDGAN with less than one-fifth of the parameters. Performing detection and enhancement simultaneously, DPNet also has much fewer parameters and higher speed in comparison with HybridDGAN. These experimental results demonstrate that DPNet performs better than HybridDGAN in speed and memory efficiency.    

\textbf{Environmental degradation performance.} Finally, we present some qualitative results, including underwater object detection and image enhancement tasks, to demonstrate the performance of our DPNet on various environmental degradation cases. 

Fig.~\ref{de1} and Fig.~\ref{de2} show examples of detection visualization on UEDB and Underwater-Cityscapes datasets. For light interference and color casts, most methods fail to complete detection, there exist errors or missed detection phenomena. Instead, DPNet completes the detection correctly. For haze-like effects, some methods and our DPNet can complete the detection task very well. For example, HybridDGAN achieves correct detection on UEDB datasets. RetinaNet, CSAM, and ERH+FERNet achieve correct detection on Underwater-Cityscapes datasets. The qualitative results show that DPNet does detection well in various environmental degradations, since the detection subnet can access the shared representation with more content information provided by the shared module. 

Fig.~\ref{en1} and Fig.~\ref{en2} show examples of enhancement visualization on UEDB and Underwater-Cityscapes datasets. For color casts and haze-like effects, we can see that DPNet recovers the fish, plate number, and buildings with a clearer edge in comparison with other enhancement methods. Similarly, DPNet can present a sharper edge line in light interference scenes. Typically, DPNet has more natural water lines instead of overexposure like ERH, FUnIE-GAN, or HybridDGAN in light interference scenes. The qualitative results show that DPNet does image enhancement well in various environmental degradations, since the enhancement subnet can access the shared representation with more structure details provided by the shared module.

\section{Conclusion}\label{con}
Underwater object detection and image enhancement are two interrelated tasks. Leveraging information coming from the two tasks can benefit each task. Based on these factual opinions, we propose a bilevel optimization formulation for jointly learning underwater object detection and image enhancement, and then unroll to a dual perception network (DPNet) for the two tasks. DPNet with one shared module and two task subnets seeks a shared representation. The shared representation contains more structural details from the detection optimization and rich content information from the enhancement optimization. Finally, we derive a cooperative training strategy to optimize parameters for DPNet. Extensive experiments on several underwater datasets demonstrate that our method outputs not only visually favoring images but also higher detection accuracy than the state-of-the-art approaches.

\section{ACKNOWLEDGEMENT}    
This work was supported in part by the National Key R\&D Program of China under Grant 2020YFB1313503; in part by the National Natural Science Foundation of China under Grants 61922019, 62027826, and 61772105; in part by the LiaoNing Revitalization Talents Program under Grant XLYC1807088; and in part by the Fundamental Research Funds for the Central Universities.

\bibliographystyle{IEEEtran}
\bibliography{egbib}

\end{document}